# AN IMPROVED TRACKING USING IMU AND VISION FUSION FOR MOBILE AUGMENTED REALITY APPLICATIONS

Kriti Kumar, Ashley Varghese, Pavan K Reddy, Narendra N, Prashanth Swamy, M Girish Chandra and Balamuralidhar P

TCS Innovation Labs, Bangalore, India

## ABSTRACT

*Mobile Augmented Reality (MAR) is becoming an important cyber-physical system application given the ubiquitous availability of mobile phones. With the need to operate in unprepared environments, accurate and robust registration and tracking has become an important research problem to solve. In fact, when MAR is used for tele-interactive applications involving large distances, say from an accident site to insurance office, tracking at both the ends is desirable and further it is essential to appropriately fuse inertial and vision sensors' data. In this paper, we present results and discuss some insights gained in marker-less tracking during the development of a prototype pertaining to an example use case related to breakdown/damage assessment of a vehicle. The novelty of this paper is in bringing together different components and modules with appropriate enhancements towards a complete working system.*

## KEYWORDS

*Augmented Reality, Inertial Measurement Unit, Sensor Fusion, CAD model, Extended Kalman Filter, Tracking,*

## 1. INTRODUCTION

Augmented Reality (AR) is a technology by which a user's view of the real world is augmented with additional information like graphics, video, and/or speech [1], [2]. By exploiting people's visual and special skills, AR brings information into user's real world as an illusion [2]. In fact, with the increasing importance, it is redefined as "AR is an approach to visualizing cyber information on top of physical imagery and manipulating cyber information through interaction with real-world objects" [3]. AR has emerged as a promising approach for visualization and interaction in multiple domains, including medical, construction, advertising, manufacturing and gaming. AR applications [4] require fast and accurate computational solutions to several complex problems, such as user and real object tracking, occlusion etc. Further, they have to be robust, degrading gracefully and recovering quickly after a failure. However, the computational complexity of the implementation should be such that they can be performed in real time.

Current generation mobile devices, such as smartphones and tablets, with an array of sophisticated sensors, sufficient processing/storage capabilities and superior network connectivity make them ideal platforms for building AR applications - Mobile AR (MAR) is the current trend. MAR due to its inherent mobility and resource constraints with the phone/tablet; exacerbate the challenges existing in conventional AR apart from bringing new ones. Some of the significant challenges remain related to sensor noise, precise localization, information fusion, complex information visualization and computational complexity [3].

This paper addresses the issue of tracking in MAR system. AR system combines virtual and physical objects to create an illusion of the virtual elements being a natural part of the real world. To achieve this illusion, the virtual and real objects have to be aligned to each other –this is called Registration. Also, it should operate in real-time; failing to do so results in inconsistencies and destroys the illusion. Registration uses the mechanism of tracking to align virtual and real content together. Tracking determines the position and orientation (pose) of the camera in all Six Degrees of Freedom (6DoF), which can either be absolute or relative to the surroundings. Hence, efficient tracking mechanisms are needed to realize such systems.

Many attempts have been made to solve the problem of object tracking in real-time where the environment is prepared in advance using fiducial markers [5], [6]. This approach puts constraints in the realization of MAR systems and calls for marker-less tracking methods, especially for outdoor applications. A few attempts have been made in this direction [6], [7], but it still remains an open research problem. Most of the object tracking techniques are vision based [8], as they are capable of accurately estimating pose with respect to the object of interest. However, they require dedicated algorithms and hardware for their high computational workload. Although being stable in the long term, they lack robustness against fast motion dynamics and occlusions.

The inertial sensors (accelerometer, gyroscope and magnetometer) that come with the mobile devices used in MAR can be additionally used for tracking [9]. Inertial Measurement Unit (IMU) based tracking is insensitive to occlusions, shadowing and provides a fast response. Therefore, they can be considered to assist the vision whenever it fails due to loss of visual features.

Some literature exists [10], [11] which talk about IMU and vision fusion for tracking in the field of AR. In [12], the authors describe an edge based tracking using a textured 3D model of buildings combined with inertial sensors to obtain accurate pose prediction under fast movement. Similar work has been reported in [13], which combines rate gyroscopes with vision for estimating motion blur which is used for tuning the feature detector in visual sensor for robust tracking. Another work [14], utilizes FAST feature detector and image patches with inertial measurements to speed up the computation of pose from two images. Most of the existing literature uses special hardware with highly sophisticated inertial sensors. However, tracking using mobile device sensors which have a limited accuracy is still a difficult problem.

This paper presents a hybrid tracking system for MAR applications which combines the IMU and vision based techniques to provide a robust tracking experience. This combination overcomes the shortcomings associated with the individual component based tracking. An Extended Kalman Filter (EKF) based algorithm combines an edge based tracker for accurate localization with fast rotational response from inertial sensors like gyroscope, accelerometer and magnetometer in the mobile device. This results in a tracking system which recovers easily from dynamic occlusions and failures. Although the individual components are well established, they need to be appropriately tuned before combining them together for the application scenario. This fusion method appears to be novel to the best of our knowledge for the application scenario described below. The main advantage of this method lies in its simplicity, ease of implementation and reproducibility.

This paper discusses the issues associated with tracking and approaches to solve them using a running example of MAR for Tele-assistance, which is a tele-inspection and assistance application designed for remote assistance for end users. This system is described in detail in Section 2. Section 3 discusses the method adopted for tracking on the mobile device side in brief. This is followed by Section 4, which presents the methodology used for pose estimation and tracking on the remote expert side, the main contribution of this paper. Section 5 presents the results and discussion for tracking on the remote expert side. Finally, Section 6 concludes the work.

## 2. OVERVIEW

This section discusses the overview of the MAR running example setup and the tracking algorithm approach.

### 2.1. System Description

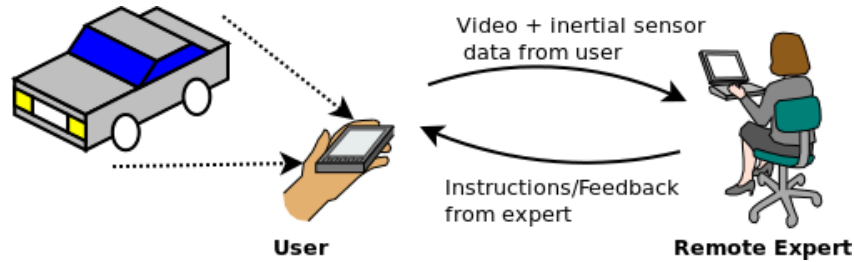

Figure 1. Application Scenario

Figure 1 shows the application scenario of a tele-interactive application for remote assistance for car drivers in case of breakdown or damage assessment. This system makes use of mobile devices, such as smartphones to offer remote support to end users by experienced supervisors at remote location. Using a mobile device application, which is installed on the user's hand-held device, video frames are transferred from the user side to the remote expert side along with the IMU data. The expert analyses the video frames and marks the object/area of interest that is transferred back to the user in the form of text, graphic or audio which gets augmented on the mobile device. This helps the user to focus on the area of interest provided by the expert and troubleshoot the problem based on instructions provided. Figure 2 shows a novel generic framework of the system which can be carried to other scenarios like healthcare, assisted maintenance, education etc. Keeping the example use case in mind, a prototype of a remote tele-interactive system is designed and its implementation is discussed in the next sub-section.

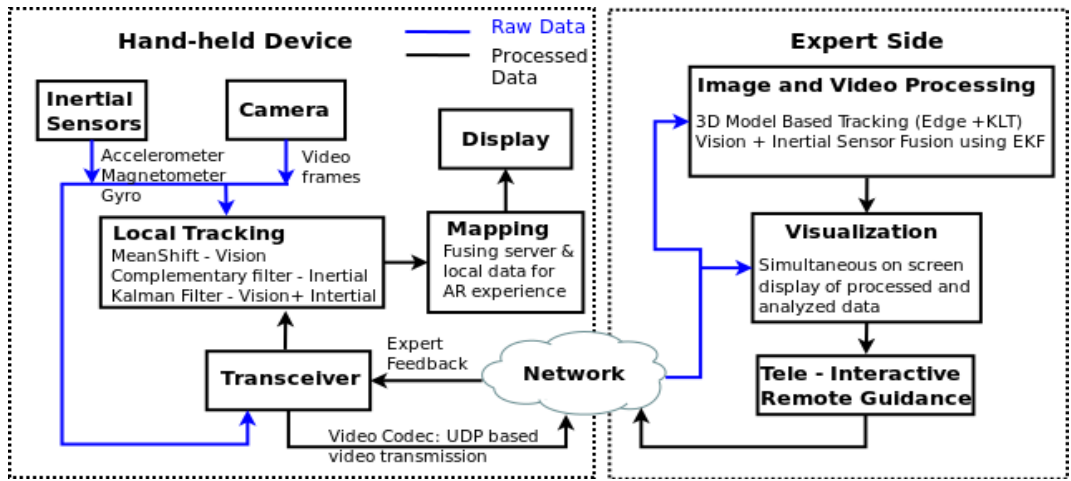

Figure 2. Generic Application Framework

The International Journal of Multimedia & Its Applications (IJMA) Vol.6, No.5, October 2014## 2.2. Prototype

The prototype implementation to test our approach was realized through an Android OS based smartphones connected through a Wifi/3G network to a computer. An Android application is developed to perform data acquisition, transmission and on-board tracking. The remote expert is equipped with a high performance computer capable of complex data processing. The video frames are compressed using MJPEG video format and transmitted via a UDP socket to facilitate real time transmission through the 3G/Wifi network. The interesting challenges and solutions associated with the communication aspect of MAR are discussed in [15]. One of the key challenges is the communication delay between the smartphone and the remote expert computer, which mandates the need for separate tracking on both the expert-side as well as the mobile device-side.

Tracking on the mobile device uses a combination of IMU and vision based methods. The overall pose of the mobile device is obtained by taking the orientation estimate from IMU and position estimate from vision using Adaptive Meanshift algorithm [16]. This pose is used to overlay a graphic on the object/area of interest using OpenGL on the mobile device screen to assist the user. More information on the implementation details of the tracking on mobile device side are discussed in Section 3. Though Adaptive Meanshift algorithm was used here, other vision based tracking techniques can also be explored. Tracking on the mobile device side using SLAM [17] is in progress.

Tracking on the expert side also uses a combination of IMU and vision based methods. Here, the vision based method utilizes a 3D Computer Aided Design (CAD) model of the object; in this case a car model. It gives the orientation and position estimate of the mobile device with respect to the object. As discussed earlier, vision based methods have several shortcomings. Therefore, there is a need to fuse IMU based tracking with vision to compute a robust pose for markerless tracking application in AR. This paper emphasizes on tracking on the remote expert side where fusion is performed using EKF and is discussed in Section 4.

## 3. A BRIEF NOTE ON MOBILE DEVICE SIDE TRACKING

Marker-less AR tracking requires an accurate estimation of the camera pose with respect to the object co-ordinates in all 6DoF. The mobile device side tracking uses the IMU sensor data together with the vision sensor data to do the pose estimation. The IMU sensor gives data with respect to their local/body co-ordinate system. To transform this into world co-ordinate system, a rotation matrix is derived from the IMU sensor data (discussed in Section 4.2).

The orientation estimate is obtained from the IMU sensor data. The inherent noise associated with IMU sensor data is filtered using a 4th order Butterworth low pass filter. The accelerometer is prone to noise and bias errors, whereas the magnetometer is affected by magnetic interference. To obtain a robust orientation estimate from these sensors along with the gyroscope, complementary filter [18] is used to fuse these sensor data due to its low computational complexity. This method overcomes the problems associated with each individual IMU sensors and results in an accurate orientation estimate.

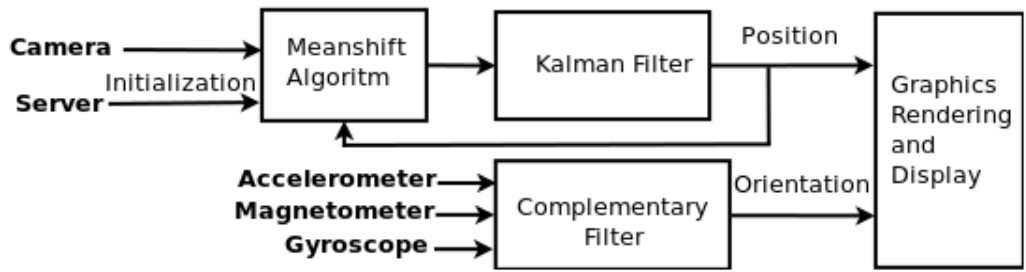

Figure 3. Hybrid Tracking Framework

The position estimation is addressed using vision based techniques alone because accelerometer data is prone to bias errors which worsen due to double integration required for position estimate. Adaptive Meanshift algorithm is used here to track the object in the image frame. This is realized by locating the object of interest in the current frame by receiving a 2D image co-ordinate from server and using colour feature to model the object of interest.

The vision based algorithm as discussed above cannot handle occlusion, and hence, a hybrid tracking method is adopted to estimate the pose as shown in Figure 3, which combines the IMU sensor and vision based tracking. This hybrid method uses the orientation from IMU sensor and position from the vision based algorithm. A Kalman filter is used for this implementation in which a constant velocity model is used to model the dynamics of the system. In normal operation, the filter works in correction mode which estimates the region of interest for meanshift tracking algorithm. During occlusion phase, the filter operates in the prediction mode following the process model. The resulting pose from the hybrid tracking algorithm is then used for defining the pose of the virtual camera, which renders the graphics on the phone screen as shown in Figure 4. This figure demonstrates the communication between the user and the remote expert side, where the expert selects the tyre of the car as his region of interest; the same is communicated to the mobile device side and a graphic gets overlayed on that region.

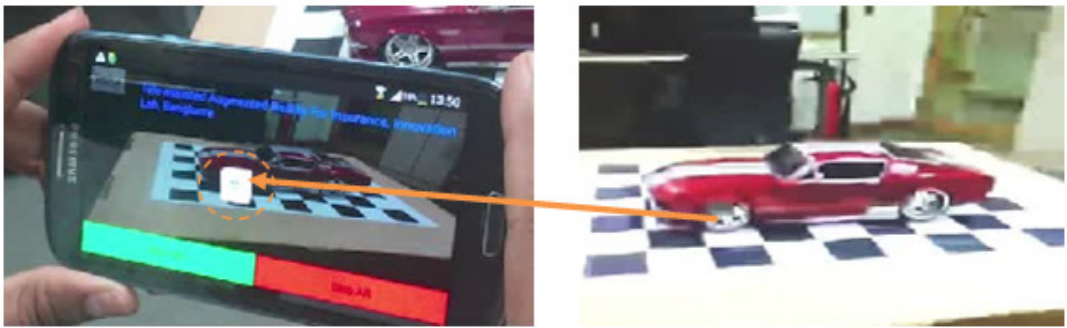

(a) Mobile device display with graphic                    (b) Real Scene
Figure 4. Tracking on Mobile Device Side

## 4. METHODOLOGY FOR REMOTE EXPERT SIDE TRACKING

This section discusses the different methods explored for tracking on the remote expert side. The methods used were, tracking using IMU sensors, tracking using vision based methods and finally tracking using the fusion of IMU and vision based methods. Tracking using fusion resulted in a more stable and robust system. In this study, quaternion is used for representing orientation and fusion is performed using EKF. A brief theoretical background explaining quaternions and EKF is presented next.

## 4.1. Quaternions and EKF

Quaternions are extensively used for representing the orientation as they offer several advantages over other conventional methods [9] like, DCM, Euler angle representation etc. They are computationally efficient and do not suffer from gimbal lock. Mathematically, quaternion is a special case of hyper complex number [19] and it can be represented as,

$$q = q_0 + q_1 i + q_2 j + q_3 k, \text{ with } i^2 = j^2 = k^2 = ijk = -1 \qquad (1)$$

where $q_0$ is the scalar component(*s*) and $q_1, q_2, q_3$ are the vector(*v*) components. It can also be represented as $q = [s, v]$. The properties of quaternion algebra are discussed in [19]. Quaternions support normal algebraic properties and operations. Despite other operations, quaternion multiplication is not commutative. Multiplication of two quaternions result in another quaternion which is represented as,

$$q_a \otimes q_b = [s_a, a][s_b, b] \qquad (2)$$
$$= [s_a s_b - a \cdot b, s_a b + s_b a + a \times b] \qquad (3)$$

where *a.b*, is the dot product and $a \times b$ is cross product of vectors *a* and *b*.

Another concept used here is of Extended Kalman Filter (EKF) which is a preferred tool for performing fusion [20]. EKF is a special case of Kalman Filter extensively used for state prediction and estimation of non-linear system model. The state equation is given as,

$$\dot{x} = f(x) + w(t) \qquad (4)$$

where *x* is the state which describes the dynamics of the system, *f* is the nonlinear transition matrix and *w* is white Gaussian process noise. EKF linearizes the process model every time instant by computing the Jacobian of transition matrix about the state vector. Standard Kalman Filter equations are now applicable on this linearized model. The state equations are given as,

$$\Delta x_{k+1} = F \Delta x_k + w_k \qquad (5)$$
$$z_k = H_k x_k + v_k \qquad (6)$$

where *F* is the state transition matrix for linearized system, $z_k$ is the measurement at time instant *k*, *H* is the measurement matrix and *v* is the measurement noise. The next section discusses the reference coordinate system and transformations involved for performing IMU and vision fusion as the data needs to be transformed in a common representation.

## 4.2. Reference Co-ordinate System

Tracking requires an accurate and robust estimate of the camera pose with respect to the object/global co-ordinates. There are four co-ordinate frames involved in the entire tracking setup as shown in Figure 5. These are:

**Camera co-ordinate frame {c}:** This frame is attached to the camera on the mobile device with its z-axis pointing along the optical axis and origin located at the camera optical center.

**IMU/Body co-ordinate frame {b}:** This frame is attached to the IMU (accelerometer, magnetometer, and gyroscope) on the mobile device.

**Object co-ordinate frame {o}:** This frame is attached to the object; in this work it is 3D CAD model of a car.

**Global/World co-ordinate frame {g}:** This is the frame in which the user is navigating and hence, the pose of the camera and IMU should be determined w.r.t this frame to help in fusing the IMU and vision based techniques for robust tracking.

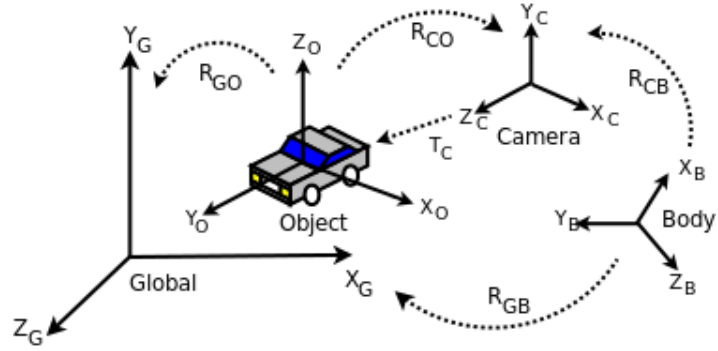

Figure 5. Reference Co-ordinate System and Transformations

IMU method (described in Section 4.3) provides orientation of the body {b} with respect to the global co-ordinate frame {g} ($R_{GB}$) and vision method (described in Section 4.4) provides orientation of the object {o} with respect to the camera co-ordinate frame {c} ($R_{CO}$). It is therefore necessary to bring both in a common frame i.e. {g} frame to perform fusion ($R_{GB}$) (described in Section 4.5). This requires the vision estimates to be transformed using the following equations.

$$R_{GB_{vision}} = R_{GO} \cdot (R_{CO})^{-1} \cdot R_{CB} \tag{7}$$

where $R_{GO}$ and $R_{CB}$ are fixed and require one time computation.

$$R_{GO} = R_{GB_{IMU}} \cdot (R_{CB})^{-1} \cdot R_{CO} \tag{8}$$

The object is always fixed w.r.t the global co-ordinates so, $R_{GO}$ is computed only once. Although both camera and IMU are fixed w.r.t the mobile device, there exists a transformation ($R_{CB}$) between {c} and {b}. $R_{CB}$ is required for transforming between the two frames to perform fusion. This calibration was done offline using InerVis toolbox for Matlab [21].

### 4.3. IMU Based Method for Orientation Estimation and Tracking

This method uses IMU available on the mobile device for calculating the orientation of the body {b} with respect to the global co-ordinate frame {g}. The rotation matrix obtained using accelerometer and magnetometer is used to compute the quaternion. It is then fused with the gyroscope data using EKF to overcome the issues associated with their independent usage. The IMU fusion framework is shown in Figure 6.

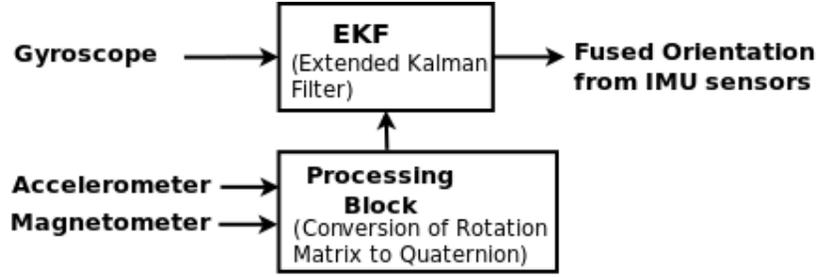

Figure 6. Block diagram of EKF for Orientation Estimation

The state vector $x$ consists of angular velocity $\omega$ and quaternion $q$ for orientation and is given as:
$$x = [\omega_x \ \omega_y \ \omega_z \ q_0 \ q_1 \ q_2 \ q_3]^T \quad (9)$$

Bias errors associated with the IMU are not considered as part of the state vector. The measurement consists of angular velocity $\omega^b$ obtained from gyroscope and quaternion $q^{gb}$ obtained from accelerometer-magnetometer combination. The measurement vector $z$ is:
$$z = [\omega_x^b \ \omega_y^b \ \omega_z^b \ q_0^{gb} \ q_1^{gb} \ q_2^{gb} \ q_3^{gb}]^T \quad (10)$$

The process model of a rigid body under rotational motion [22] is used as it worked well for this application. The equations used are:
$$\dot{\omega} = \frac{-1}{\tau}\omega \quad (11)$$
$$\dot{q} = \frac{1}{2}q \otimes \omega \quad (12)$$

where $\tau$ is the time constant of the process model and $\otimes$ is the quaternion multiplication given in (2). The linearized transition matrix $F_Q$ is:

$$F_Q = \begin{bmatrix} e^{-\delta t/\tau_1} & 0 & 0 & 0 & 0 & 0 & 0 \\ 0 & e^{-\delta t/\tau_2} & 0 & 0 & 0 & 0 & 0 \\ 0 & 0 & e^{-\delta t/\tau_3} & 0 & 0 & 0 & 0 \\ \frac{-\hat{x}_5 \delta t}{2} & \frac{-\hat{x}_6 \delta t}{2} & \frac{-\hat{x}_7 \delta t}{2} & 1 & \frac{-\hat{x}_1 \delta t}{2} & \frac{-\hat{x}_2 \delta t}{2} & \frac{-\hat{x}_3 \delta t}{2} \\ \frac{\hat{x}_4 \delta t}{2} & \frac{-\hat{x}_7 \delta t}{2} & \frac{\hat{x}_6 \delta t}{2} & \frac{\hat{x}_1 \delta t}{2} & 1 & \frac{\hat{x}_3 \delta t}{2} & \frac{-\hat{x}_2 \delta t}{2} \\ \frac{\hat{x}_7 \delta t}{2} & \frac{\hat{x}_4 \delta t}{2} & \frac{-\hat{x}_5 \delta t}{2} & \frac{\hat{x}_2 \delta t}{2} & \frac{-\hat{x}_3 \delta t}{2} & 1 & \frac{\hat{x}_1 \delta t}{2} \\ \frac{-\hat{x}_6 \delta t}{2} & \frac{\hat{x}_5 \delta t}{2} & \frac{\hat{x}_4 \delta t}{2} & \frac{\hat{x}_3 \delta t}{2} & \frac{\hat{x}_2 \delta t}{2} & \frac{-\hat{x}_1 \delta t}{2} & 1 \end{bmatrix} \quad (13)$$

where $\hat{x}_i$ is the $i^{th}$ component of state vector $x$ and $\delta t$ is the sampling interval of IMU sensor.

Although this model is extensively used in literature [23], the process noise covariance $Q_Q$ and measurement noise covariance $R_Q$ are appropriately tuned for this application after elaborate study.

$Q_Q = \begin{bmatrix} q_{11} & 0_{3 \times 4} \\ 0_{4 \times 3} & q_{22} \end{bmatrix}$ where $q_{11} = I_3 (D/2\tau)(1 - \exp(-2\delta t/\tau))$ and $q_{22} = I_4 q_p$ are the process noise taken for gyroscope and quaternion respectively. $D$ is the variance of the white noise process.

$R_Q = \begin{bmatrix} \text{var}_\omega & 0_{3 \times 4} \\ 0_{4 \times 3} & \text{var}_{q^{gb}} \end{bmatrix}$ where $\text{var}_\omega = I_3 \sigma_{\omega^b}^2$ and $\text{var}_{q^{gb}} = I_4 \sigma_{q^{gb}}^2$ are the measurement noise taken for gyroscope and quaternion respectively. The measurement noise is assumed white Gaussian with zero mean. The values of the filter tuning parameters are given in Table I. The observation matrix $H_Q = I_7$ where $I_n$ is an identity matrix dimension of $n$. This fusion results in a robust orientation estimate from IMU which overcomes the drift problem of gyroscope and noise associated with accelerometer and magnetometer.

The International Journal of Multimedia & Its Applications (IJMA) Vol.6, No.5, October 2014

## 4.4. Vision Based Method for Pose Estimation and Tracking

3D vision based tracking method tracks the position and orientation of the mobile device camera relative to the referenced object in real time. In our application, the object of interest for tracking is a car. For effective tracking, any vision algorithm needs to have strong and stable features to track. Vehicles generally exhibit large homogeneous regions with strong edge features near the boundaries between different parts. Due to this, it is appropriate to track such objects based on edge features. Here, we use a combination of 3D CAD model based edge tracking and KLT (Kanade Lucas Tomasi) point tracking to estimate the pose of the vehicle for each frame. Figure 7 shows the block diagram of the tracking method adopted.

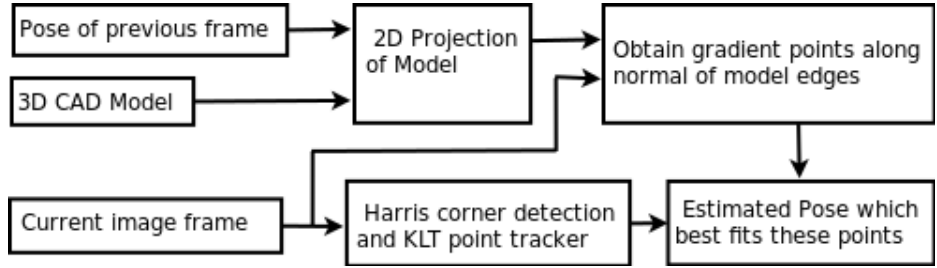

Figure 7. Block diagram of Model Based Tracking

The 3D wireframe model, which represents the actual measurements of the object, is assumed to be known. Since the algorithm tracks the object incrementally from previous position, it is necessary to initialize the tracking by providing the pose of the object for the first frame. The initialization is done using the Dementhon approach [24], which returns the pose of the object with respect to the camera based on point correspondence between the model and 2D projection image.

In order to track the object in successive frames, the hybrid tracking algorithm provided by [25] is used. During each successive frame, the algorithm projects the wireframe 3D model onto 2D, based on the pose obtained from previous frame. The algorithm then searches for a strong image gradient along the normal direction of these projected edges as shown in Figure 8(b). Within the area bounded by the wire-frame model, keypoints are obtained and tracked using KLT tracker. With the assumption that the keypoints lie on the surface of the polygon defined by the model, an optimal pose is estimated which best fits these points. The optimal pose is obtained by minimizing the function given by,

$$\arg\min_{^{c}M_o} \sum_i \left(p_i - pr\left(^{c}M_o \cdot P_0\right)\right)^2 \qquad (14)$$

where $p_i \in R^2$ are the points obtained from image gradient, $P_o \in R^3$ represents the corresponding points on the model, $^{c}M_o$ is the extrinsic matrix representing the transformation from camera to the object and $pr()$ indicates the projection from 3D to 2D using the intrinsic matrix. The intrinsic matrix parameters are obtained offline by calibrating the camera.

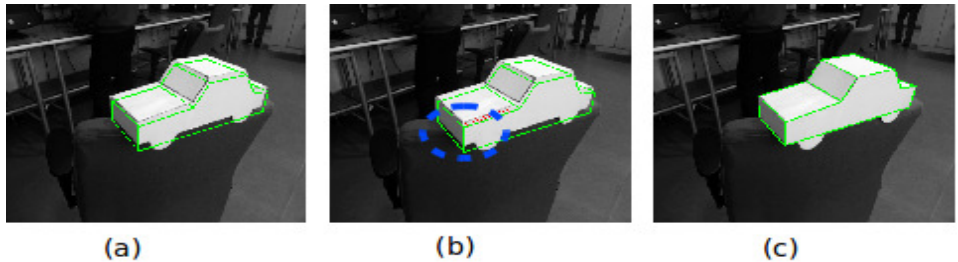

Figure 8. Various stages involved in Tracking

Figure 8 illustrates the various stages of tracking for an image frame. Figure 8(a) shows the wireframe model projected on the current frame with the pose obtained for previous frame. Figure 8(b) shows the error between the previous pose and edge location in current frame, and Figure 8(c) shows model fitting after pose estimation for the current frame.

**4.5. IMU and Vision Based Method for Pose Estimation and Tracking**

As discussed above, vision based method is more accurate than IMU in most cases, but fails to handle occlusion, fast motion etc. In such cases, IMU method can assist vision method in recovering from failure, thereby providing robust tracking [26]. This method uses both IMU and vision based techniques, fusing them in an appropriate way for robust pose estimation. Both these techniques provide the orientation estimate reliably to some extent as discussed earlier. Position however, is addressed by vision method alone since computing it using IMU requires motion constraints to be followed and is not covered in the scope of this work.

IMU and vision fusion is performed using a single EKF, incorporating the process model for orientation (discussed in Section 4.3) with the constant velocity model [12] for position. The fused EKF computes the overall pose of the mobile device with respect to {g} frame. This fused EKF as shown in Figure 9 was specially designed and thoroughly tested for our application scenario.

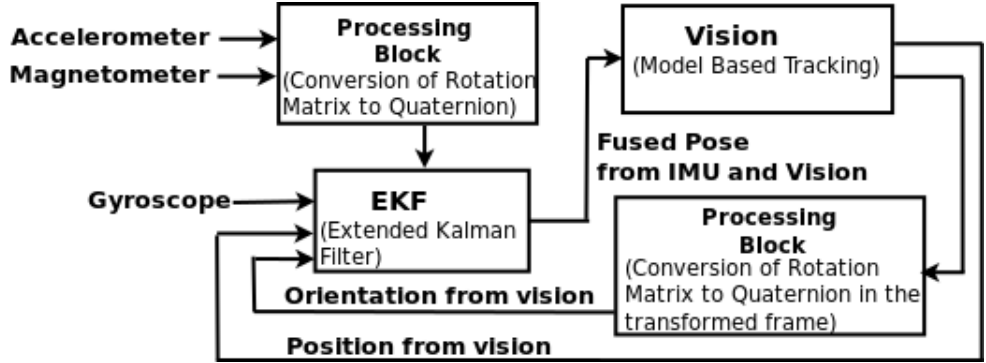

Figure 9. Block diagram of Single EKF for Pose Estimation

The state vector $x$ of the fused filter consists of angular velocity $\omega$ quaternion for orientation $q$ position $t$ and velocity $v$.

$$x = [\omega_x \ \omega_y \ \omega_z \ q_0 \ q_1 \ q_2 \ q_3 \ t_x \ t_y \ t_z \ v_x \ v_y \ v_z]^T \quad (15)$$

The measurement vector $z$ consists of angular velocity $\omega^b$ from gyroscope, quaternion $q^{gb}$ from accelerometer-magnetometer and quaternion $q^{gbi}$ and position $t^o$ from vision. The quaternion $q^{gbi}$ is the orientation estimate from vision when transformed to the {g} frame using (7&8) for performing fusion.

$$z = [\omega_x^b \ \omega_y^b \ \omega_z^b \ q_0^{gb} \ q_1^{gb} \ q_2^{gb} \ q_3^{gb} \ q_0^{gbi} \ q_1^{gbi} \ q_2^{gbi} \ q_3^{gbi} \ t_x^o \ t_y^o \ t_z^o]^T \quad (16)$$

The fused linearized transition matrix $F_{QT}$ for the use case considered here is:

$$F_{QT} = \begin{bmatrix} F_Q & 0_{7\times 6} \\ 0_{6\times 7} & F_T \end{bmatrix} \quad (17)$$

where $F_Q$ is as defined in Section 4.3 and $F_T$ is:

$$F_T = \begin{bmatrix} 1 & 0 & 0 & \delta t_I & 0 & 0 \\ 0 & 1 & 0 & 0 & \delta t_I & 0 \\ 0 & 0 & 1 & 0 & 0 & \delta t_I \\ 0 & 0 & 0 & 1 & 0 & 0 \\ 0 & 0 & 0 & 0 & 1 & 0 \\ 0 & 0 & 0 & 0 & 0 & 1 \end{bmatrix} \quad (18)$$

where $\delta t_I$ is the sampling time between image frames. The process noise covariance $Q_{QT}$ is:

$$Q_{QT} = \begin{bmatrix} Q_Q & 0_{7\times 6} \\ 0_{6\times 7} & Q_T \end{bmatrix} \quad (19)$$

where $Q_Q$ is as defined in the Section 4.3 and $Q_T$ is:

$$Q_T = \begin{bmatrix} \frac{\delta t_I^4}{4} & 0 & 0 & \frac{\delta t_I^3}{2} & 0 & 0 \\ 0 & \frac{\delta t_I^4}{4} & 0 & 0 & \frac{\delta t_I^3}{2} & 0 \\ 0 & 0 & \frac{\delta t_I^4}{4} & 0 & 0 & \frac{\delta t_I^3}{2} \\ \frac{\delta t_I^3}{2} & 0 & 0 & \delta t_I^2 & 0 & 0 \\ 0 & \frac{\delta t_I^3}{2} & 0 & 0 & \delta t_I^2 & 0 \\ 0 & 0 & \frac{\delta t_I^3}{2} & 0 & 0 & \delta t_I^2 \end{bmatrix} * K \quad (20)$$

The measurement noise covariance $R_{QT}$ is:

$$R_{QT} = \begin{bmatrix} R_Q & 0_{7\times 4} & 0_{7\times 3} \\ 0_{4\times 7} & R_I & 0_{4\times 3} \\ 0_{3\times 7} & 0_{3\times 4} & R_T \end{bmatrix} \quad (21)$$

where $R_I = I_4 \sigma^2_{q^{gbi}}$ is the measurement noise of the quaternion from vision, $R_T = I_3 \sigma^2_{T^0}$ is the measurement noise of the position from vision and $R_Q$ is the measurement noise of the quaternion from IMU having values as discussed in the Section 4.3.

The observation matrix $H_{QT}$ is:

$$H_{QT} = \begin{bmatrix} H_Q & 0_{7\times 3} & 0_{7\times 3} \\ 0_{4\times 3} & H_I & 0_{4\times 6} \\ 0_{3\times 3} & 0_{3\times 4} & H_T \end{bmatrix} \quad (22)$$

where $H_I = I_4$ is the observation matrix for quaternion estimate from vision. $H_T = [I_3 \ \ 0_3]$ is the observation matrix for position estimate from vision and $H_Q$ is the observation matrix for the quaternion from IMU as discussed in the Section 4.3. All the measurement noise is assumed white Gaussian with zero mean and covariance values as given in Table 1. The results of the fused implementation are discussed in the next section.

Table 1. Filter Tuning Parameters and their Values.

| Parameter | Symbol | Value |
|---|---|---|
| Process noise for Quaternion | $q_p$ | 0.01 |
| Time constant of Process model for Orientation | $\tau$ | 0.5s |
| Sampling interval of IMU sensor | $\delta_t$ | 0.02s |
| Variance of white noise process | D | 0.4 rad$^2$/s$^2$ |
| Measurement noise variance for Gyroscope | $\sigma^2_{\omega^b}$ | 0.1 rad$^2$/s$^2$ |
| Measurement noise variance for Quaternion from IMU | $\sigma^2_{q^{gb}}$ | 0.001 |
| Sampling interval between image frames | $\delta_{tI}$ | 0.0333s |
| Constant | K | 250 |
| Measurement noise variance for Quaternion from vision | $\sigma^2_{q^{gbi}}$ | 0.0001 |
| Measurement noise variance for Position from vision | $\sigma^2_{T^0}$ | 0.001 m$^2$ |

## 5. RESULTS AND DISCUSSION

An Android application was developed to demonstrate the tracking performance of the proposed fusion filter. Samsung Galaxy S3 with Quad-core 1.4 GHz Cortex-A9 processor and 8MP back camera was used as handheld device for data acquisition and tracking. The remote system was a Linux based system. IMU data was acquired at 50Hz, and appropriate filtering and calibration was done. Video frames were acquired with a resolution of 640x480 at 30fps. In order to provide synchronization between the IMU and video frames acquisition for fusion, IMU data was sent as part of the header of each video frame. An initial estimate of the user position was also sent using GPS for tracking initialization. The fused EKF was tuned with parameters given in Table I which were obtained after extensive experimentation and testing for our application. The tracking results obtained during the development of end to end prototype are discussed next.

The proposed fusion filter for marker-less tracking on the expert side was implemented on hardware and tested for two Cases: Movement with Occlusion and Movement without Occlusion. Figure 10 depicts the flowchart of the fused EKF implementation for these two cases.

### 5.1. Movement without Occlusion

In this case, the fusion filter was made to operate in prediction and correction mode continuously using measurements obtained from both the IMU and vision. Since vision based tracking was more accurate for slow movement than IMU, the measurement error covariance of IMU was taken higher than that of vision for orientation estimate. Figure 11(a-c) shows the plots of orientation obtained for this case. This figure shows the filtered result to be smooth and closely following the vision measurements than the IMU which was obvious as per the filter setting. Figure 11(d-f) shows the position obtained from vision and the filtered estimate. The tracking performed well and was validated through visual inspection.

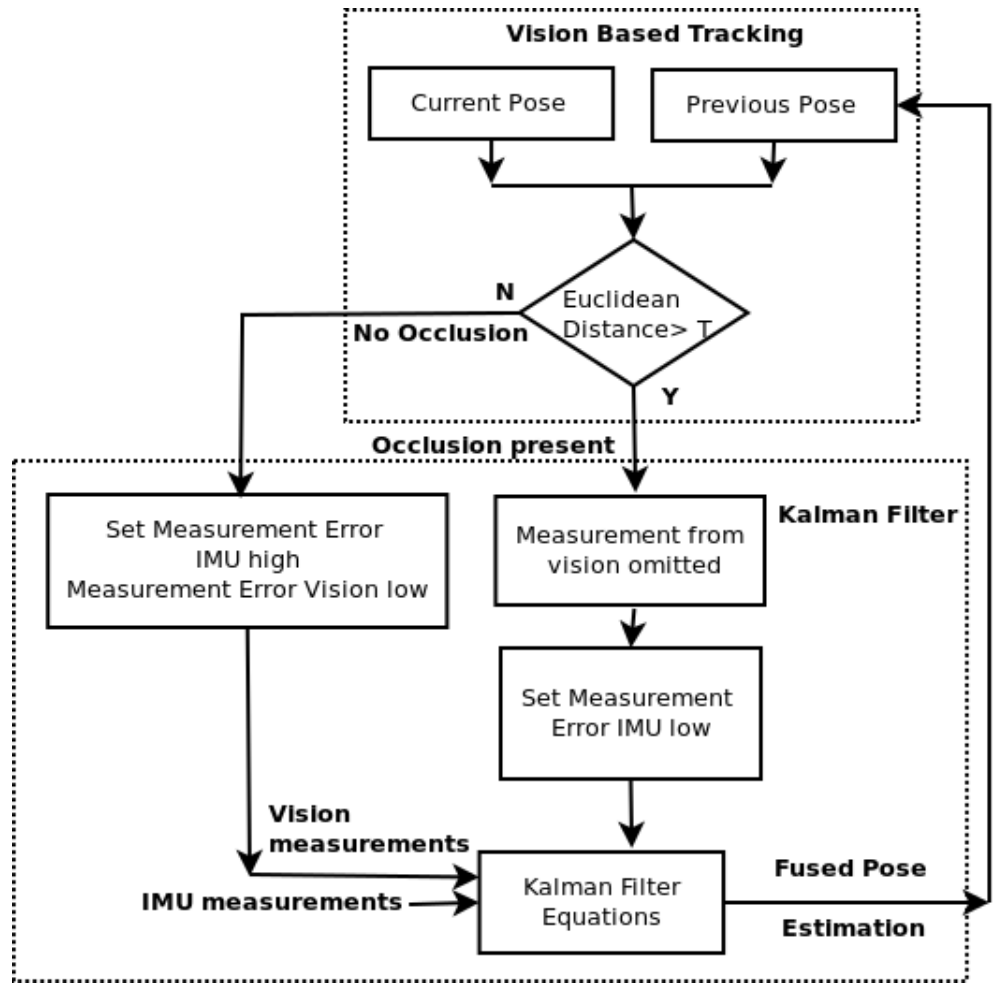

Figure 10. Flowchart of the Fused System for Occlusion handling

## 5.2. Movement with Occlusion

The proposed filter was capable of handling short term occlusions. Whenever, the vision method failed to track around the predicted location provided by the fusion filter, the object was assumed to be hidden. In the current implementation, we observed that occlusion caused significant change in the position estimate from vision. Therefore, the object was assumed to be occluded whenever, the distance between the current and last estimated states of position crossed a fixed defined threshold T (as given in Figure 10) which was derived empirically.

Once occlusion was detected, the position and orientation measurement estimate from vision were not considered. The filter state was reset with the orientation provided by the IMU and it was made to operate in prediction mode for estimating the position. Also, the measurement error covariance of IMU was reduced to allow the filter to closely follow the IMU measurements. When occlusion phase was over, measurement error covariance values were reset to the normal motion case. Figure 12 demonstrates the occlusion handling feature of the fusion filter. Figure 13(a-f) shows the plots of orientation and position for the occlusion case. As shown in the plots, whenever occlusion occurred the vision estimates went wrong (shown as spikes in red), at these instants the vision algorithm was reset with IMU estimates and the fused EKF was made to closely follow them.

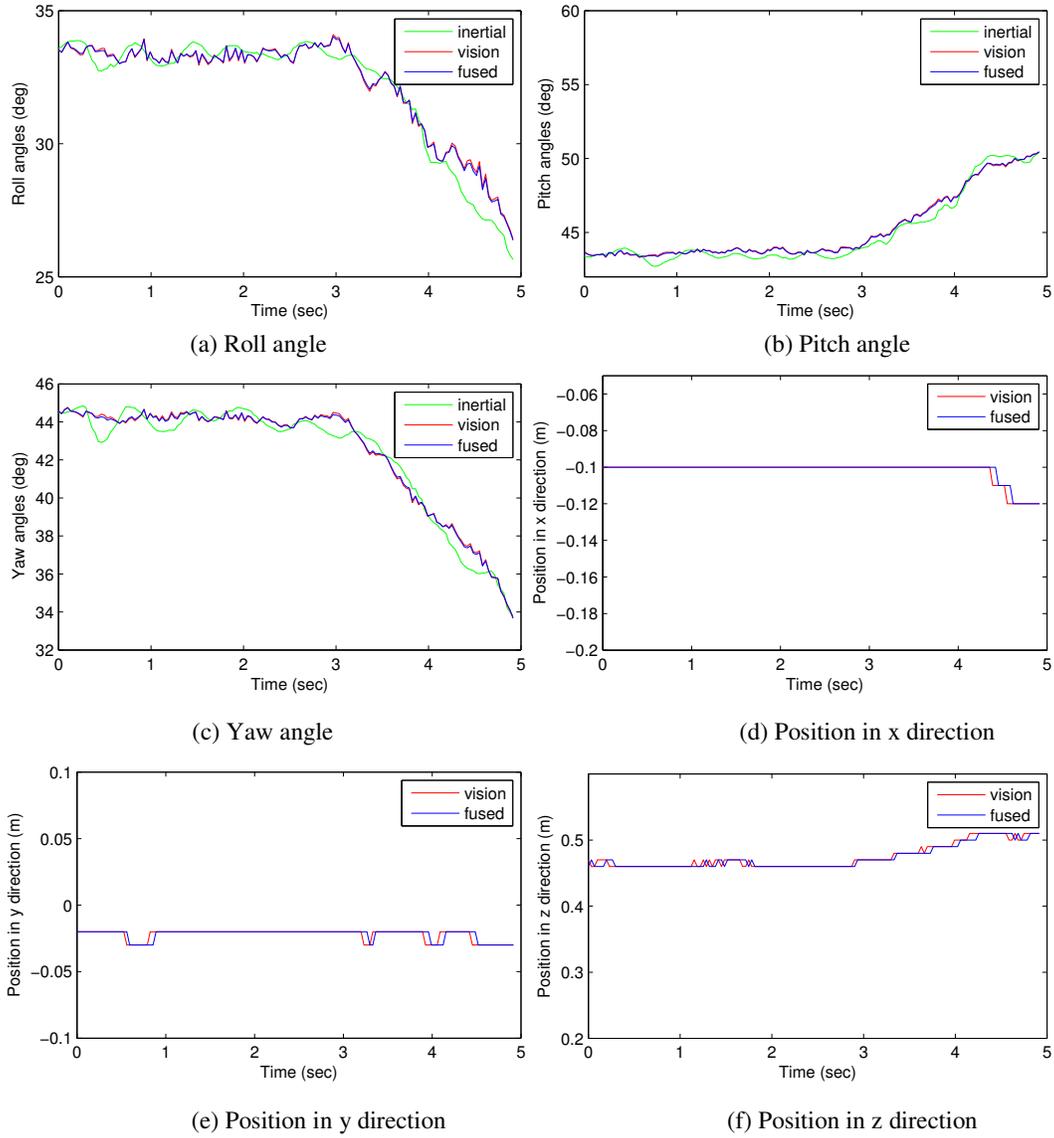

Figure 11. Tracking Results of the Fused System without Occlusion

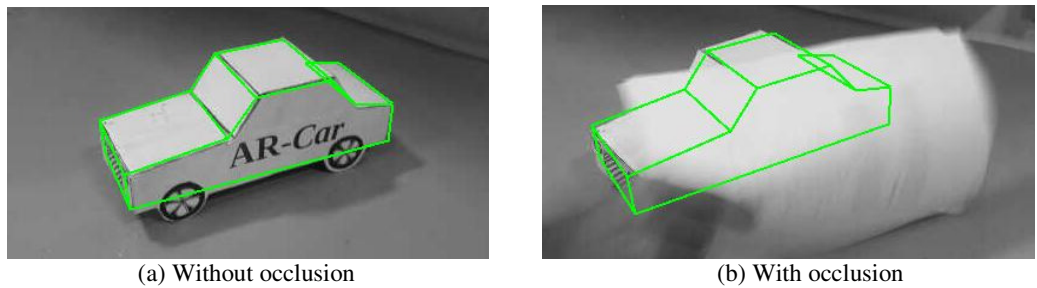

Figure 12. Tracking Results

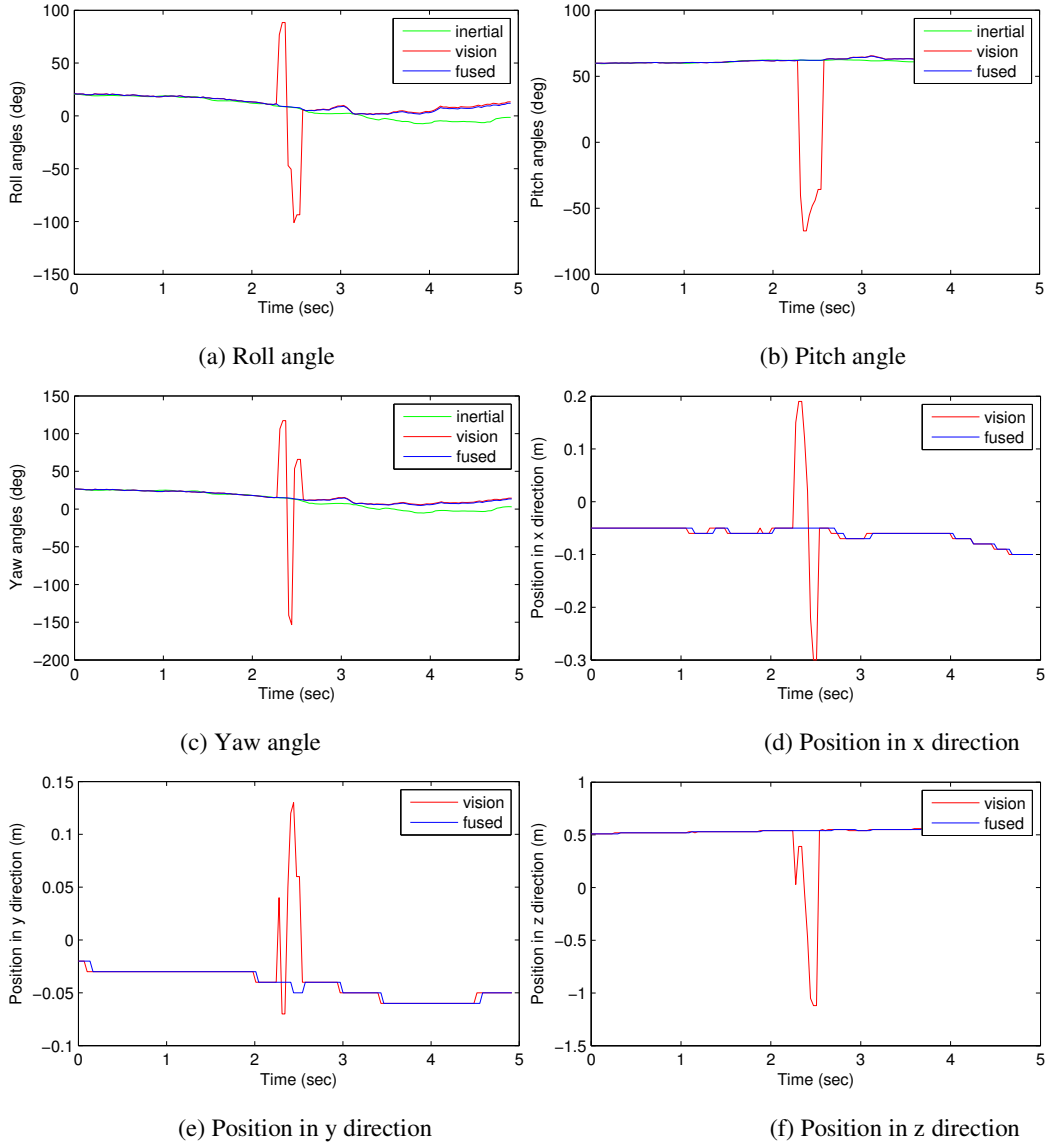

(a) Roll angle

(b) Pitch angle

(c) Yaw angle

(d) Position in x direction

(e) Position in y direction

(f) Position in z direction

Figure 13. Tracking Results of the Fused System with Occlusion

## 6. CONCLUSION

This paper presents an improved tracking approach using fusion of IMU and vision for Markerless AR applications. An end to end solution of the overall system implementation for the example use case is provided. The proposed tracking approach worked well in real time, fusing the fast rotational response from IMU with accurate pose from vision resulting in a robust tracking solution. The results demonstrated the potential of the fused approach in handling dynamic failures, like occlusion.

Although the tracking framework adopted worked well for our application, issues related to 3D position estimation using IMU still needs to be addressed. This can be considered next to further enhance the tracking especially in case of fast motion. The initialization of vision based tracking, for demonstration purpose, was done manually. It can be streamlined by extracting the pose of the vehicle automatically from a single monocular image. Further, the dependency on having the CAD model apriori can be eliminated by using deformable CAD models where a single generic model can be deformed to fit any kind of vehicle. Although a primitive car model with strong edge features has been used here, there is already work in progress towards considering an augmented set of features including edge points, colour, information from other feature generator like Kaze [27] towards applying the methodology and techniques to the more realistic scenario while maintaining a comparable accuracy. Here, we have shown a use case which involves tele-assistance for vehicle breakdown/damage assessment. However, similar framework can be adopted for any other applications on mobile devices like health care, maintenance and education which require tele-assistance.